\documentclass[11pt,a4paper]{article}
\usepackage[hyperref]{naaclhlt2019}
\usepackage{times}
\usepackage{latexsym}
\usepackage{sidecap}

\usepackage{url}

\usepackage[utf8]{inputenc} 
\usepackage[T1]{fontenc}    
\usepackage{url}            
\usepackage{booktabs}       
\usepackage{amsfonts}       
\usepackage{nicefrac}       
\usepackage{microtype}      
\usepackage{amsmath}
\usepackage{graphicx}
\usepackage{booktabs}
\usepackage{enumitem}
\usepackage{dblfloatfix}

\DeclareMathOperator{\sgn}{sgn}

\usepackage{colortbl}

\newcommand{\abs}[1]{\left| #1 \right|}

\usepackage{algorithm}
\usepackage{algpseudocode}
\usepackage{array}
\usepackage{booktabs}
\usepackage{multirow}
\usepackage{pbox}
\usepackage{caption}
\usepackage{wrapfig}

\usepackage{listings}
\usepackage{xargs}   
\usepackage{xcolor}  

\newcommand\Tstrut{\rule{0pt}{2.0ex}}       
\newcommand\Bstrut{\rule[-0.8ex]{0pt}{0pt}} 
\newcommand{\TBstrut}{\Tstrut\Bstrut} 

\usepackage[colorinlistoftodos,prependcaption,textsize=tiny]{todonotes}
\newcommandx{\unsure}[2][1=]{\todo[linecolor=red,backgroundcolor=red!25,bordercolor=red,#1]{#2}}
\newcommandx{\change}[2][1=]{\todo[linecolor=blue,backgroundcolor=blue!25,bordercolor=blue,#1]{#2}}
\newcommandx{\info}[2][1=]{\todo[linecolor=OliveGreen,backgroundcolor=OliveGreen!25,bordercolor=OliveGreen,#1]{#2}}
\newcommandx{\improvement}[2][1=]{\todo[linecolor=Plum,backgroundcolor=Plum!25,bordercolor=Plum,#1]{#2}}
\newcommandx{\thiswillnotshow}[2][1=]{\todo[disable,#1]{#2}}

\newcommand{\cmmnt}[1]{\ignorespaces}

\usepackage{etoolbox}

\makeatletter
\expandafter\patchcmd\csname\string\algorithmic\endcsname{\itemsep\z@}{\itemsep=0.2ex plus2pt}{}{}
\makeatother

\algrenewcommand\algorithmicindent{0.9em}%

\makeatletter
\newcommand{\algmargin}{\the\ALG@thistlm}
\makeatother
\newlength{\whilewidth}
\algdef{SE}[parWHILE]{parWhile}{EndparWhile}[1]
  {\parbox[t]{\dimexpr\linewidth-\algmargin}{%
     \hangindent\whilewidth\strut\algorithmicwhile\ #1\ \algorithmicdo\strut}}{\algorithmicend\ \algorithmicwhile}%
\algnewcommand{\parState}[1]{\State%
  \parbox[t]{\dimexpr\linewidth-\algmargin}{\strut #1\strut}}

\algnewcommand\algorithmicinput{\textbf{Procedure:}}
\algnewcommand\algorithmicoutput{\textbf{Output:}}
\algnewcommand\Input{\item[\algorithmicinput]}%
\algnewcommand\Output{\item[\algorithmicoutput]}%

\aclfinalcopy 
\def\aclpaperid{4010} 
	
\title{Active Learning for New Domains in Natural Language Understanding}

\author{Stanislav Peshterliev, John Kearney, Abhyuday Jagannatha, Imre Kiss, Spyros Matsoukas\\
Alexa Machine Learning, Amazon.com \\
\texttt{\{stanislp,jkearn,abhyudj,ikiss,matsouka\}@amazon.com}
 \\}

\date{\vspace{-5em}}

\begin{document}
\maketitle%
\begin{abstract}
We explore active learning~(AL) for improving the accuracy of new domains in a natural language understanding~(NLU) system. We propose an algorithm called Majority-CRF that uses an ensemble of classification models to guide the selection of relevant utterances, as well as a sequence labeling model to help prioritize informative examples. Experiments with three domains show that Majority-CRF achieves 6.6\%-9\% relative error rate reduction compared to random sampling with the same annotation budget, and statistically significant improvements compared to other AL approaches. Additionally, case studies with human-in-the-loop AL on six new domains show 4.6\%-9\% improvement on an existing NLU system.
\end{abstract}

\section{Introduction}
\label{sec:intro}

Intelligent voice assistants (IVA) such as Amazon Alexa, Apple Siri, Google Assistant, and Microsoft Cortana, are becoming increasingly popular. For IVA, natural language understanding (NLU) is a main component~\cite{de2008spoken}, in conjunction with automatic speech recognition (ASR) and dialog management (DM). ASR converts user's speech to text. Then, the text is passed to NLU for classifying the action or ``intent'' that the user wants to invoke (e.g.,\ PlayMusicIntent, TurnOnIntent, BuyItemIntent) and recognizing named-entities (e.g.,\ Artist, Genre, City). Based on the NLU output, DM decides the appropriate response, which could be starting a song playback or turning off lights. NLU systems for IVA support functionality in a wide range of domains, such as music, weather, and traffic. Also, an important requirement is the ability to add support for new domains.

The NLU models for Intent Classification (IC) and Named Entity Recognition (NER) use machine learning to recognize variation in natural language. Diverse, annotated training data collected from IVA users, or ``annotated live utterances,'' are essential for these models to achieve good performance.
As such, new domains frequently exhibit suboptimal performance due to a lack of annotated live utterances. While an initial training dataset can be bootstrapped using grammar generated utterances and crowdsourced collection (Amazon Mechanical Turk), the performance that can be achieved using these approaches is limited because of the unexpected discrepancies between anticipated and live usage. Thus, a mechanism is required to select live utterances to be manually annotated for enriching the training dataset.


Random sampling is a common method for selecting live utterances for annotation.  However, in an IVA setting with many users, the number of available live utterances is vast. Meanwhile, due to the high cost of manual annotation, only a small percentage of utterances can be annotated. As such, in a random sample of live data, the number of utterances relevant to new domains may be small. Moreover, those utterances may not be informative, where informative utterances are those that, if annotated and added to the training data, reduce the error rates of the NLU system.
Thus, for new domains, we want a sampling procedure which selects utterances that are both relevant and informative.

Active learning (AL)~\cite{settles.tr09} refers to machine learning methods that can interact with the sampling procedure and guide the selection of data for annotation.
In this work, we explore using AL for live utterance selection for new domains in NLU. Authors have successfully applied AL techniques to NLU systems with little annotated data overall~\cite{tur2003active, Shen:2004:MAL:1218955.1219030}.
The difference with our work is that, to the best of our knowledge, there is little published AL research that focuses on data selection explicitly targeting new domains.

We compare the efficacy of least-confidence~\cite{lewis1994heterogeneous} and query-by-committee~\cite{freund1997selective} AL for new domains.
Moreover, we propose an AL algorithm called Majority-CRF, designed to improve both IC and NER of an NLU system. Majority-CRF uses an ensemble of classification models to guide the selection of relevant utterances, as well as a sequence labeling model to help prioritize informative examples. Simulation experiments on three different new domains show that Majority-CRF achieves 6.6\%-9\% relative improvements in-domain compared to random sampling, as well as significant improvements compared to other active learning approaches. 

\section{Related Work}
\label{sec:related} 

Expected model change~\cite{settles2008multiple} and expected error reduction~\cite{roy2001toward} are AL approaches based on decision theory. Expected model change tries to select utterances that cause the greatest change on the model. Similarly, expected error reduction tries to select utterances that are going to maximally reduce generalization error. Both methods provide sophisticated ways for ascertaining the value of annotating an utterance. However, they require computing an expectation across all possible ways to label the utterance, which is computationally expensive for NER and IC models with many labels and millions of parameters. Instead, approaches to AL for NLU generally require finding a proxy, such as model uncertainty, to estimate the value of getting specific points annotated.
 
Tur et al.\ studied least-confidence and query-by-committee disagreement AL approaches for reducing the annotation effort~\cite{tur2005combining, tur2003active}. Both performed better than random sampling, and the authors concluded that the overall annotation effort could be halved.
We investigate both of these approaches, but also a variety of new algorithms that build upon these basic ideas.

Schutze et al.~\cite{schutze2006performance} showed that AL is susceptible to the missed cluster effect when selection focuses only on low confidence examples around the existing decision boundary, missing important clusters of data that receive high confidence. They conclude that AL may produce a sub-optimal classifier compared to random sampling with a large budget. To solve this problem Osugi et al.~\cite{osugi2005balancing} proposed an AL algorithm that can balance exploitation (sampling around the decision boundary) and exploration (random sampling) by reallocating the sampling budget between the two. In our setting, we start with a representative seed dataset, then we iteratively select and annotate small batches of data that are used as feedback in subsequent selections, such that extensive exploration is not required. 

To improve AL, Hong-Kwang and Vaibhava~\cite{kuo2005active} proposed to exploit the similarity between instances. Their results show improvements over simple confidence-based selection for data sizes of less than 5,000 utterances. A computational limitation of the approach is that it requires computing the pairwise utterance similarity, an $\mathcal{O}(N^2)$ operation that is slow for millions of utterances available in production IVA. However, their approach could be potentially sped-up with techniques like locality-sensitive hashing.


\section{Active Learning For New Domains}
\label{sec:approach}

We first discuss random sampling baselines and standard active learning approaches. Then, we describe the Majority-CRF algorithm and the other AL algorithms that we tested.

\subsection{Random Sampling Baselines}

A common strategy to select live utterances for annotation is random sampling. We consider two baselines: uniform random sampling and domain random sampling.

Uniform random sampling is widespread because it provides unbiased samples of the live utterance distribution. However, the samples contain fewer utterances for new domains because of their low usage frequency. Thus, under a limited annotation budget, accuracy improvements on new domains are limited.

Domain random sampling uses the predicted NLU domain to provide samples of live utterances more relevant to the target domains. However, this approach does not select the most informative utterances. 

\subsection{Active Learning Baselines}

AL algorithms can select relevant and informative utterances for annotation. Two popular AL approaches are least-confidence and query-by-committee.

\textit{Least-confidence}~\cite{lewis1994heterogeneous} involves processing live data with the NLU models and prioritizing selection of the utterances with the least confidence. The intuition is that utterances with low confidence are difficult, and ``teaching'' the models how they should be labeled is informative. However, a weakness of this method is that out-of-domain or irrelevant utterances are likely to be selected due to low confidence. This weakness can be alleviated by looking at instances with medium confidence using measures such as least margin between the top-$n$ hypotheses~\cite{scheffer2001active} or highest Shannon entropy \cite{settles2008analysis}.

\textit{Query-by-committee (QBC)}~\cite{freund1997selective} uses different classifiers (e.g.,\ SVMs, MaxEnt, Random Forests) that are trained on the existing annotated data. Each classifier is applied independently to every candidate and the utterances assigned the most diverse labels are prioritized for annotation. One problem with this approach is that, depending on the model and the size of the committee, it could be computationally expensive to apply on large datasets. 

\subsection{Majority-CRF Algorithm}
\label{sec:alg}

Majority-CRF is a confidence-based AL algorithm that uses models trained on the available NLU training set but does not rely on predictions from the full NLU system. Its simplicity compared to a full NLU system offers several advantages. First, fast incremental training with the selected annotated data. Second, fast predictions on millions of utterances. Third, the selected data is not biased to the current NLU models, which makes our approach reusable even if the models change.

Algorithm~\ref{al_alg} shows a generic AL procedure that we use to implement Majority-CRF, as well as other AL algorithms  that we tested. We train an ensemble of models on positive data from the target domain of interest (e.g.,\ Books) and negative data that is everything not in the target domain (e.g., Music, Videos). Then, we use the models to filter and prioritize a batch of utterances for annotation. After the batch is annotated, we retrain the models with the new data and repeat the process.

To alleviate the tendency of the least-confidence approaches to select irrelevant data, we add unsupported utterances and sentence fragments to the negative class training data of the AL models. This helps keep noisy utterances on the negative side of the decision boundary, so that they can be eliminated during filtering. Note that, when targeting several domains at a time, we run the selection procedure independently and then deduplicate the utterances before sending them for annotation.

\begin{algorithm}

\caption{Generic AL procedure that selects data for a target domain}
\label{al_alg}

\begin{algorithmic}[1]
\Require
\Statex $D\gets$ positive and negative training data
\Statex $P\gets$ pool of unannotated live utterances
\Statex $i\gets$ iterations, $m\gets$ mini-batch size
\Ensure
\Statex $\{\mathcal{M}^k\} \gets$ set of selection models
\Statex $\mathcal{F}\gets$ filtering function
\Statex $\mathcal{S}\gets$ scoring function
\Input
\Repeat{ $i$ iterations}
\parState {Train selection models $\{\mathcal{M}^k\}$ on $D$}
\parState {$\forall$ $x_i \in P$ obtain prediction scores $y^k_i$ = $\mathcal{M}^k(x_i)$ }
\parState {$P' \gets$ \{$x_i \in P \ : \mathcal{F}(y^0_i .. y^k_i)$ \} }
\parState {$C \gets$ \{$x_i \in P' : $  $m$ with the smallest score $\mathcal{S}(y^0_i .. y^k_i)$\} } 
\parState {Send $C$ for manual annotation}
\parState {After annotation is done $D\gets D \cup C$ and $P\gets P \setminus C$}
\Until

\end{algorithmic}

\end{algorithm} 

\begin{SCtable*}

\centering

{\footnotesize


\renewcommand{\arraystretch}{1.5}

\begin{tabular}{| l | c | c | c |  }

\hline

\rowcolor[gray]{.91}

Algorithm & Models $\{\mathcal{M}^i\}$ & Filter $\mathcal{F}$ & Scoring $\mathcal{S}$ \TBstrut\\

\specialrule{1.0pt}{0.5pt}{0.5pt}

AL-Logistic & lg & $\sgn(y^{lg}) > 0$ & $y^{lg}$ \TBstrut\\
QBC-SA & lg, sq, hg & $\sum \sgn(y^k) \in \{-1, 1\}$ & $\sum \abs{y^k}$ \TBstrut\\
QBC-AS & lg, sq, hg & $\sum \sgn(y^k) \in \{-1, 1\}$ & $\abs{\sum y^k}$ \TBstrut\\
Majority-SA & lg, sq, hg & $\sum \sgn(y^k) > 0$ & $\sum \abs{y^k}$ \TBstrut\\
Majority-AS & lg, sq, hg & $\sum \sgn(y^k) > 0$ & $\abs{\sum y^k}$ \TBstrut\\

QBC-CRF & lg, sq, hg, CRF & $\sum \sgn(y^k) \in \{-1, 1\}$ & $p^{lg} \times p^{crf}$ \TBstrut\\
\textbf{Majority-CRF} & lg, sq, hg, CRF & $\sum \sgn(y^k) > 0$ & $p^{lg} \times p^{crf}$ \TBstrut\\

\hline

\end{tabular}


\caption{\label{algorithm-details} AL algorithms evaluated. $lg$, $sq$, $hg$ refer to binary classifiers (committee members) trained with logistic, squared and hinge loss functions, respectively. $y^i$ denotes the score of committee member $i$, $p^{crf}$ denotes the confidence of the CRF model and $p^{lg} = (1 + e^{- y^{lg}})^{-1}$ denotes the confidence of the logistic classifier. In all cases, we prioritize by smallest score $\mathcal{S}$.}

}

\end{SCtable*}

\textbf{Models.} We experimented with $n$-gram linear binary classifiers trained to minimize different loss functions: $\mathcal{M}^{lg}\gets$ logistic, $\mathcal{M}^{hg}\gets$ hinge, and $\mathcal{M}^{sq}\gets$ squared. Each classifier is trained to distinguish between positive and negative data and learns a different decision boundary. Note that we use the raw unnormalized prediction scores $\{y^{lg}_i,y^{hg}_i,y^{sq}_i\}$ (no sigmoid applied) that can be interpreted as distances between the utterance $x_i$ and the classifiers decision boundaries at $y = 0$. The classifiers are implemented in Vowpal Wabbit~\cite{JL07a} with $\{1, 2, 3\}$-gram features. To directly target the NER task, we used an additional $\mathcal{M}^{cf}\gets$ CRF, trained on the NER labels of the target domain. 

\textbf{Filtering function.} We experimented with $\mathcal{F}^{maj} \gets \sum \sgn(y^k) > 0$, i.e., keep only majority positive prediction from the binary classifiers, and $\mathcal{F}^{dis} \gets \sum \sgn(y^k) \in \{-1, 1\}$, i.e., keep only prediction where there is at least one disagreement.

\textbf{Scoring function.} When the set of models $\{\mathcal{M}^k\}$ consists of only binary classifiers, we combine the classifier scores using either the sum of absolutes $\mathcal{S}^{sa}\gets \sum |y_i^{k}|$ or the absolute sum $\mathcal{S}^{as}\gets |\sum y_i^{k}|$. $\mathcal{S}^{sa}$ prioritizes utterances where all scores are small (i.e., close to all decision boundaries), and $\mathcal{S}^{as}$ prioritizes utterances where either all scores are small or there is large disagreement between classifiers (e.g., one score is large negative, another is large positive, and the third is small). Both $\mathcal{S}^{sa}$ and $\mathcal{S}^{as}$ can be seen as generalization of least-confidence to a committee of classifiers. When the set of models $\{\mathcal{M}^k\}$ includes a CRF  model $\mathcal{M}^{cf}$, we compute the score with $\mathcal{S}^{cg}\gets P_{cf}(i) \times P_{lg}(i) $, i.e., the CRF probability $P_{cf}(i)$ multiplied by the logistic classifier probability $P_{lg}(i) = \sigma(y^{lg}_i)$, where $\sigma$ is the sigmoid function. Note that we ignore the outputs of the squared and hinge classifiers for scoring, though they are still be used for filtering. 

The full set of configurations we evaluated is given in Table~\ref{algorithm-details}, which specifies the choice of parameters $\{\mathcal{M}^k\}, \mathcal{F}, \mathcal{S}$ used in Algorithm~\ref{al_alg}.

AL-Logistic and QBC serve as baseline AL algorithms. The QBC-CRF and Majority-CRF models combine the IC focused binary classifier scores with the NER focused sequence labeling scores and use filtering by disagreement and majority (respectively) to select informative utterances. To the best of our knowledge, this is a novel architecture for active learning in NLU.

Mamitsuka et al.~\cite{mamitsuka1998query} proposed bagging to build classifier committees for AL. Bagging refers to random sampling with replacement of the original training data to create diverse classifiers. We experimented with bagging but found that it is not better than using different classifiers. 

\section{Experimental Results}
\label{sec:experiments}

\subsection{Evaluation Metrics}

We use Slot Error Rate (SER)~\cite{Makhoul99performancemeasures}, including the intent as slot, to evaluate the overall predictive performance of the NLU models. SER as the ratio of the number of slot prediction errors to the total number of reference slots. Errors are insertions, substitutions and deletions.  We treat the intent misclassifications as substitution errors.

\subsection{Simulated Active Learning}

AL requires manual annotations which are costly. Therefore, to conduct multiple controlled experiments with different selection algorithms, we simulated AL by taking a subset of the available annotated training data as the unannotated candidate pool, and ``hiding'' the annotations.
As such, the NLU system and AL algorithm had a small pool of annotated utterances for simulated ``new'' domains.
Then, the AL algorithm was allowed to choose relevant utterances from the simulated candidate pool. Once an utterance is selected, its annotation is revealed to the AL algorithm, as well as to the full NLU system.

\textbf{Dataset.} We conducted experiments using an internal test dataset of 750K randomly sampled live utterances, and a \cmmnt{38,329,255} training dataset of 42M utterances containing a combination of grammar generated and randomly sampled live utterances. The dataset covers 24 domains, including Music, Shopping, Local Search, Sports, Books, Cinema and Calendar.

\textbf{NLU System.} Our NLU system has one set of IC and NER models per domain. The IC model predicts one of its in-domain intents or a special out-of-domain intent which helps with domain classification. The IC and NER predictions are ranked into a single n-best list based on model confidences~\cite{su2018re}. We use MaxEnt~\cite{berger1996maximum} models for IC and the CRF models for NER~\cite{lafferty2001conditional}.


\textbf{Experimental Design.} We split the training data into a 12M utterances initial training set for IC and NER, and a 30M utterance candidate pool for selection. We choose Books, Local Search, and Cinema as target domains to simulate the AL algorithms, see Table~\ref{dataset-description}. Each target domain had 550-650K utterances in the candidate pool. The rest of the 21 non-target domains have 28.5M utterances in the candidate pool. We also added 100K sentence fragments and out-of-domain utterances to the candidate pool, which allows us to compare the susceptibility of different algorithms to noisy or irrelevant data. This experimental setup attempts to simulate the production IVA use case where the candidate pool has a large proportion of utterances that belong to different domains.

We employed the different AL algorithms to select 12K utterances per domain from the candidate pool, for a total 36K utterance annotation budget.
Also, we evaluated uniform (Rand-Uniform) and domain (Rand-Domain) random sampling with the same total budget.
We ran each AL configuration twice and average the SER scores to account for fluctuations in selection caused by the stochasticity in model training. For random sampling, we ran each selection five times.

\begin{table}[!tb]
\centering
{\footnotesize
\vspace{-10px}
\begin{tabular}{ | l | c | c | l | }
\hline
\rowcolor[gray]{.91}
\pbox[c][12pt][c]{\textwidth}{Domain} & Train & Test & Examples \\ \hline

Books & \pbox[c][25pt][c]{\textwidth}{\cmmnt{833620} 290K} & {13K \cmmnt{13205}} & \pbox[c][31pt][c]{100pt}{
``search in mystery books'' \\
``read me a book''
} \\ \hline

\pbox[c][12pt][c]{\textwidth}{Local\\ Search} & \pbox[c][25pt][c]{\textwidth}{\cmmnt{267636} 260K} & {16K \cmmnt{16178}} & \pbox[c][31pt][c]{100pt}{
``mexican food nearby'' \\
``pick the top bank''
} \\ \hline

\pbox[c][12pt][c]{\textwidth}{Cinema} & \pbox[c][25pt][c]{\textwidth}{\cmmnt{269,675} 270K} & {9K \cmmnt{9051}} & \pbox[c][31pt][c]{100pt}{
``more about hulk'' \\
``what's playing in theaters''
} \\ \hline
\end{tabular}%
\caption{\label{dataset-description} Simulated ''new`` target domains for AL experiments. The target domain initial training datasets are 90\% grammar generated data. The other 21 ''non-new`` domains have on average 550k initial training datasets with 60\% grammar generated data and 40\% live data.}
}
\end{table}

\begin{table*}[!tb]
{\footnotesize
\centering
\resizebox{\textwidth}{!}{%
\begin{tabular}{| l | l | l | c c | c  c | c c | c |}
\hline 

\multirow{2}{*}{Algorithm Group} & \multirow{2}{*}{Algorithm ($i=6$)} & Overall & \multicolumn{2}{c |}{Books} & \multicolumn{2}{c | }{Local Search} & \multicolumn{2}{c |}{Cinema} & Non-Target \TBstrut\\

& & \#Utt & \#Utt & \pbox[c][12pt][c]{\textwidth}{$\Delta$SER} & \#Utt & \pbox[c][12pt][c]{\textwidth}{$\Delta$SER} & \#Utt & \pbox[c][12pt][c]{\textwidth}{$\Delta$SER} & \#Utt \TBstrut\\
\specialrule{1.0pt}{0.5pt}{0.5pt}

\multirow{2}{*}{Random} 
  & \pbox[c][11pt][c]{\textwidth}{Rand-Uniform} & 35.8K & 747 & 1.20 & 672 & 3.37 & 547 & 0.57 & 33.8K \TBstrut\\
  & \pbox[c][11pt][c]{\textwidth}{Rand-Domain} & 35.7K & 9853 & 1.52 & 9453 & 4.23 & 9541 & 1.75 & 06.8K \TBstrut\\
\hline

\multirow{3}{*}{Single Model} 
  & \pbox[c][11pt][c]{\textwidth}{AL-Logistic($i$=1)} & 34.9K & 5405 & 4.76 & 7092 & 6.54 & 5224 & 6.09 & 17.1K \TBstrut\\
  & \pbox[c][11pt][c]{\textwidth}{AL-Logistic} & 35.1k & 5524 & 6.77 & 7709 & 7.24 & 5330 & 7.29 & 16.5K \TBstrut\\  
\hline

\multirow{4}{*}{Committee Models}
 & \pbox[c][11pt][c]{\textwidth}{QBC-AS} & 35.0K & 4768 & 7.18 & 7869 & 8.57 & 4706 & 8.72 & 17.6K \TBstrut\\
 & \pbox[c][11pt][c]{\textwidth}{QBC-SA} & 35.0K & 4705 & 7.12 & 7721 & 8.96 & 4790 & 7.52 & 17.7K \TBstrut\\
 & \pbox[c][11pt][c]{\textwidth}{Majority-AS} & 35.1K & 5389 & \underline{7.66} & 8013 & 9.07 & 5526 & 8.98 & 16.1K \TBstrut\\
 & \pbox[c][11pt][c]{\textwidth}{Majority-SA} & 35.1K & 5267 & 7.35 & 8196 & 8.46 & 5193 & 8.42 & 16.4K \TBstrut\\

\hline

\multirow{2}{*}{Committee and CRF}
 & \pbox[c][11pt][c]{\textwidth}{QBC-CRF} & 35.1K & 3653 & 7.44 & 6593 &  \underline{9.78} & 4064 & \underline{10.26} & 20.7K \TBstrut\\
 & \pbox[c][11pt][c]{\textwidth}{Majority-CRF} & 35.1K & 6541 & \bf 8.42 & 8552 & \bf 9.92 & 6951 & \bf{11.05} & 13.0K \TBstrut\\
\hline
\end{tabular}%
}
\caption{\label{simulation-results} Simulation experimental results with 36K annotation budget. $\Delta$SER is \% relative reduction is SER compared to the initial model: Books SER 30.59, Local Search SER 39.09, Cinema SER 38.71. Higher $\Delta$SER is better. The best result is in bold, and the second best is underlined. The $i = 1$ means selection in a single iteration, otherwise if not specified selection is in six iterations ($i = 6$). Overall \#Utt shows the remaining from the 36K selected after removing the sentence fragments and out-of-domain utterances. Both target and non-target domains IC and NER models are re-retrained with the new data.}
}
\end{table*}

\subsubsection{Simulated Active Learning Results} 

Table~\ref{simulation-results} shows the experimental results for the target domains Books, Local Search, and Cinema. For each experiment, we add all AL selected data (in- and out-of-domain), and evaluate SER for the full NLU system.

We test for statistically significant improvements using the Wilcoxon test~\cite{hollander2013nonparametric} with 1000 bootstrap resamples and p-value < 0.05.

\textbf{Random Baselines.} As expected, Rand-Uniform selected few relevant utterances for the target domains due to their low frequency in the candidate pool. Rand-Domain selects relevant utterances for the target domains, achieving statistically significant SER improvements compared to Rand-Uniform. However, the overall gains are small, around 1\% relative per target domain. A significant factor for Rand-Domain's limited improvement is that it tends to capture frequently-occurring utterances that the NLU models can already recognize without errors. As such, all AL configurations achieved statistically significant SER gains compared to the random baselines.

\textbf{Single Model Algorithms.} AL-Logistic, which carries out a single iteration of confidence-based selection, exhibits a statistically significant reduction in SER relative to Rand-Domain. Moreover, using six iterations (i.e., $i$=6) further reduced SER by a statistically significant 1\%-2\% relative to AL-Logistic($i$=1), and resulted in the selection of 200 fewer unsupported utterances. This result demonstrates the importance of incremental selection for iteratively refining the selection model.

\textbf{Committee Algorithms.} AL algorithms incorporating a committee of models outperformed those based on single models by a statistically significant 1-2\% $\Delta$SER. The \textit{majority} algorithms performed slightly better than the QBC algorithms and were able to collect more in-domain utterances. The absolute sum scoring function $\mathcal{S}^{as}$ performed slightly better than the sum of absolutes $\mathcal{S}^{sa}$ for both QBC and Majority. Amongst all committee algorithms, Majority-AS performed best, but the differences with the other committee algorithms are not statistically significant.

\textbf{Committee and CRF Algorithms.} AL algorithms incorporating a CRF model tended to outperform purely classification-based approaches, indicating the importance of specifically targeting the NER task.
The Majority-CRF algorithm achieves a statistically significant SER improvement of 1-2\% compared to Majority-AS (the best configuration without the CRF). Again, the disagreement-based QBC-CRF algorithm performed worse that the majority algorithm across target domains. This difference was statistically significant on Books, but not on Cinema and Local Search.

In summary, AL yields more rapid improvements not only by selecting utterances relevant to the target domain but also by trying to select the most informative utterances. For instance, although the AL algorithms selected 40-50\% false positive utterances from non-target domains, whereas Rand-Domain selected only around 20\% false positives, the AL algorithms still outperformed Rand-Domain. This indicates that labeling ambiguous false positives helps resolve existing confusions between domains. Another important observation is that majority filtering $\mathcal{F}^{maj}$ performs better than QBC disagreement filtering $\mathcal{F}^{dis}$ across all of our experiments. A possible reason for this is that majority filtering selects a better balance of boundary utterances for classification and in-domain utterances for NER. Finally, the Majority-CRF results show that incorporating the CRF model improves the performance of the committee algorithms. We assume this is because incorporation of a CRF-based confidence directly targets the NER task.

\subsection{Human-in-the-loop Active Learning}

We also performed AL for six new NLU domains with human-in-the-loop annotators and live user data. We used the Majority-SA configuration for simplicity in these case studies. We ran the AL selection for 5-10 iterations with varying batch sizes between 1000-2000.

\begin{table}[!hbtp]
{\footnotesize
\centering
\begin{tabular}{| l | c | c | c |  }
\hline
\rowcolor[gray]{.91}
 Domain &  $\Delta$SER & \#Utt Selected & \#Utt Testset \TBstrut\\
\specialrule{1.0pt}{0.5pt}{0.5pt}
Recipes & 8.97 & 24.1K & 4.7K \TBstrut\\
\hline
LiveTV & 6.92  & 11.6K & 1.8K  \TBstrut\\
\hline 
OpeningHours & 7.05 & 6.8K & 583 \TBstrut\\
\hline
Navigation & 4.67 & 6.7K & 6.4K \TBstrut\\
\hline
DropIn & 9.00 & 5.3K & 7.2K \TBstrut\\
\hline
Membership & 7.13  & 4.2K  & 702 \TBstrut\\
\hline
\end{tabular}%
\caption{\label{human-results} AL with human annotator results. $\Delta$SER is \% relative gain compared to the existing model. Higher is better.}
}
\end{table}

Table~\ref{human-results} shows the results from AL with human annotators. On each feature, AL improved our existing NLU model by a statistically significant 4.6\%-9\%. On average 25\% of utterances are false positive. This is lower than the 50\% in the simulation because the initial training data exhibits more examples of the negative class. Around 10\% of the AL selected data is lost due to being unactionable or out-of-domain, similar to the frequency with which these utterances are collected by random sampling.

While working with human annotators on new domains, we observed two challenges that impact the improvements from AL. First, annotators make more mistakes on AL selected utterances as they are more ambiguous. Second, new domains may have a limited amount of test data, so the impact of AL cannot be fully measured. Currently, we address the annotation mistakes with manual data clean up and transformations, but further research is needed to develop an automated solution. To improve the coverage of the test dataset for new domains we are exploring test data selection using stratified sampling.

\section{Conclusions}
\label{sec:Conclusion}

In this work, we focused on AL methods designed to select live data for manual annotation. The difference with prior work on AL is that we specifically target new domains in NLU. Our proposed Majority-CRF algorithm leads to statistically significant performance gains over standard AL and random sampling methods while working with a limited annotation budget. In simulations, our Majority-CRF algorithm showed an improvement of 6.6\%-9\% SER relative gain compared to random sampling, as well as improvements over other AL algorithms with the same annotation budget. Similarly, results with live annotators show statistically significant improvements of 4.6\%-9\%  compared to the existing NLU system.

\bibliography{al_new_domains}
\bibliographystyle{acl_natbib}

\end{document}